\documentclass[graybox]{svmult}

\usepackage{mathptmx}       % selects Times Roman as basic font
\usepackage{helvet}         % selects Helvetica as sans-serif font
\usepackage{courier}        % selects Courier as typewriter font
\usepackage{makeidx}         % allows index generation
\usepackage{graphicx}        % standard LaTeX graphics tool
\usepackage{multicol}        % used for the two-column index
\usepackage[bottom]{footmisc}% places footnotes at page bottom

\usepackage{tikz}
\usepackage{multirow}

\def\checkmark{\tikz\fill[scale=0.3](0,.35) -- (.25,0) -- (1,.7) -- (.25,.15) -- cycle;}

\newcommand*\rot{\rotatebox{90}}

\newlength\lengtha 
\newlength\lengthb 

\makeindex             % used for the subject index
\begin{document}

\title*{Evaluation of Interactive\\Machine Learning Systems}
\titlerunning{Evaluation of Int. ML Systems}

\author{Nadia Boukhelifa \and Anastasia Bezerianos \and Evelyne Lutton}
\authorrunning{N. Boukhelifa et al.}

\institute{N. Boukhelifa, E. Lutton \at INRA, Universit\'{e} Paris-Saclay, 1 av. Br\'{e}tigni\`{e}res, 78850, Thiverval-Grignon, France, \email{\{nadia.boukhelifa, evelyne.lutton\}@inra.fr \\A. Bezerianos \at Univ Paris-Sud \&CNRS (LRI), INRIA, Universit\'{e} Paris-Saclay, France, e-mail:anab@lri.fr}
}
\maketitle

\abstract{
The evaluation of interactive machine learning systems remains a difficult task. These systems learn from and adapt to the human, but at the same time, the human receives feedback and adapts to the system. Getting a clear understanding of these subtle mechanisms of co-operation and co-adaptation is challenging. In this chapter, we report on our experience in designing and evaluating various interactive machine learning applications from different domains. We argue for coupling two types of validation: \emph{algorithm-centered} analysis, to study the computational behaviour of the system; and \emph{human-centered} evaluation, to observe the utility and effectiveness of the application for end-users. We use a visual analytics application for guided search, built using an interactive evolutionary approach, as an exemplar of our work. We argue that human-centered design\index{human-centered design} and evaluation\index{evaluation} complement algorithmic analysis, and can play an important role in addressing the ``black-box" effect of machine learning. Finally, we discuss research opportunities that require human-computer interaction methodologies, in order to support both the visible and hidden roles that humans play in interactive machine learning.}

\section{Introduction}
In interactive Machine Learning\index{interactive machine learning} (iML), a human operator and a machine collaborate to achieve a task, whether this is to classify or cluster a set of data points~\cite{Amershi2012,Brown2012}, to find interesting data projections~\cite{Behrisch2014,Boukhelifa2017,Ticona2012}, or to design creative art works~\cite{Lutton2005b,Song2013}. The underlying assumption is that the human-machine co-operation yields better results than a fully automated or manual system. An interactive machine learning system comprises an automated service, a user interface, and a learning component. A human interacts with the automated component via the user interface, and provides iterative feedback to a learning algorithm. This feedback may be explicit, or inferred from human behaviour and interactions. Likewise, the system may provide implicit or explicit feedback to communicate its status and the knowledge it has learnt.

The interactive approach to machine learning is appealing for many reasons including: 
\begin{itemize}
\item to integrate valuable experts knowledge that may be hard to encode directly into mathematical or computational models.
\item to help resolve existing uncertainties as a result of, for example, bias and error that may arise from automatic machine learning. 
\item to build trust by making humans involved in the modelling or learning processes. 
\item to cater for individual human differences and subjective assessments such as in art and creative applications.
\end{itemize}

Recent work in interactive machine learning has focused on developing working prototypes, but less on methods to evaluate iML systems and their various components. The question of how to effectively evaluate such systems is challenging. Indeed, human-in-the-loop approaches to machine learning bring forth not only numerous intelligibility and usability issues, but also open questions with respect to the evaluation of the various facets of the iML system, both as separate components and as a holistic entity~\cite{sacha2016}. Holzinger~\cite{Holzinger2016} argued that conducting methodically correct experiments and evaluations is difficult, time-consuming, and hard to replicate due to the subjective nature of the ``human agents" involved. Cortellessa and Cesta~\cite{cortellessa2006} found that the quantitative evaluation of mixed-initiative systems tend to focus either on problem-solving performance of the human and what they call the artificial solver, or the quality of interaction looking at user requirements and judgment of the system. This statement also applies to iML systems, where current evaluations tend to be either \emph{algorithm-centered} to study the computational behaviour of the system, or \emph{human-centered} focusing on the utility and effectiveness of the application for end-users~\cite{Boukhelifa2017,Boukhelifa2016,Boukhelifa2015}.

The aim of this chapter is to review existing evaluation methods for iML systems\index{System Review}, and to reflect upon our own experience in designing and evaluating such applications over a number of years~\cite{Bach2012,Boukhelifa2013,Collet2005,Legrand2007,Lutton2005b,Tonda2013,Valigiani2006}. The chapter is organised as follows: First we provide a review of recent work on the evaluation of iML systems focusing on types of human and system feedback, and the evaluation methods and metrics deployed in these studies. We then illustrate our evaluation method through a case study on an interactive machine learning system for guided visual search, covering both algorithm-centered and human-centered evaluations. Finally, we discuss research opportunities requiring human-computer interaction methodologies in order to support both the visible and hidden roles that humans play in machine learning.

\section{Related Work}

In this section, we review recent work that evaluates interactive machine learning systems. We consider both qualitative and quantitative evaluations. Our aim is not to provide an exhaustive survey, but rather to illustrate the broad range of existing methods and evaluation metrics.

\begin{table}[]
\centering
\begin{tabular}{@{} l
                @{\hspace*{\lengtha}}c
                @{\hspace*{\lengtha}}c
                @{\hspace*{\lengtha}}c
                @{\hspace*{\lengtha}}c
                @{\hspace*{\lengthb}}c
                @{\hspace*{\lengtha}}c
                @{\hspace*{\lengtha}}c
                @{\hspace*{\lengthb}}c
                @{\hspace*{\lengtha}}c
                @{\hspace*{\lengtha}}c
                @{\hspace*{\lengtha}}c
                @{\hspace*{\lengthb}}c
                @{\hspace*{\lengtha}}c @{}}
Paper    & \rot{Classification}  & \rot{Clustering}  & \rot{Density estimation}  & \rot{Dimensionality Reduction} & \rot{Implicit User Feedback} & \rot{Explicit User Feedback} & \rot{System Feedback} & \rot{Case Study} & \rot{User Study} & \rot{Observational Study} & \rot{Survey} & \rot{Objective Metrics} & \rot{Subjective Metrics} \\\hline
Co-integration~\cite{Azuan2017}	& \checkmark &  &  &            & \checkmark        &                   &                     &                     & \checkmark          &                        &                        & \checkmark             &                        \\
DDLite~\cite{Ehrenberg2016}	& \checkmark &  &  & &        & \checkmark                  &         \checkmark            & \checkmark          &                     &                        &                        & \checkmark             & \checkmark             \\
Interest Driven Navigation~\cite{Healey2012}	& \checkmark &  &  &  & \checkmark        & \checkmark        &         \checkmark            & \checkmark          &                     &                        &                        &                        & \checkmark             \\
ISSE~\cite{Bryan2014}	& \checkmark &  &  &             &                   & \checkmark        &                     &                     & \checkmark          &                        & \checkmark             & \checkmark             & \checkmark             \\
RCLens~\cite{Lin2017}	& \checkmark &  &  &               &                   & \checkmark        &         \checkmark            & \checkmark          &                     &                        & \checkmark             & \checkmark             & \checkmark             \\
ReGroup~\cite{Amershi2012}	& \checkmark &  &  &          & \checkmark        &                   &                     &                     & \checkmark          &                        & \checkmark             & \checkmark             & \checkmark             \\
View Space Explorer \cite{Behrisch2014}	& \checkmark &  &  &         &                   & \checkmark        &         \checkmark            & \checkmark          &                     &                        &                        & \checkmark             &             \\
Visual Classifier~\cite{Heimerl2012}	& \checkmark &  &  &           &                   & \checkmark        &         \checkmark            &                     & \checkmark          &                        & \checkmark             & \checkmark             & \checkmark 	\\
OLI~\cite{Wenskovitch2017}	&  & \checkmark &  & \checkmark      & \checkmark        &                   &                     & \checkmark          &                     &                        &                        & \checkmark                       &             \\
ForceSPIRE~\cite{Endert2012a}	&  & \checkmark &  &          & \checkmark        &                   &                     & \checkmark          &                     &                        &                        &\checkmark                        &             \\
ForceSPIRE~\cite{Endert2012b}	&  & \checkmark &  &   & \checkmark        &                   &                     &                     &                     & \checkmark             &                        & \checkmark             & \checkmark             \\
RugbyVAST~\cite{Legg2013}	&  & \checkmark &  &             &                   & \checkmark        &         \checkmark            & \checkmark          & \checkmark          &                        &                        & \checkmark             & \checkmark             \\
3D Model Repository Explorator~\cite{Gao2014}	&  & \checkmark &  &               &                   & \checkmark        &                    & \checkmark          & \checkmark          &                        &                        & \checkmark             &                        \\
User Interaction Model~\cite{Dabek2016}	&  & \checkmark &  &             & \checkmark        &                   &         \checkmark            &                     & \checkmark          &                        & \checkmark             & \checkmark             & \checkmark             \\
SelPh~\cite{Koyama2016}	& & & \checkmark & \checkmark           & \checkmark        & \checkmark        &         \checkmark            &                     & \checkmark          &                        & \checkmark             & \checkmark                       & \checkmark             \\
EvoGraphDice~\cite{Boukhelifa2017}	&  &  &  & \checkmark    & \checkmark        & \checkmark        &         \checkmark            &                     & \checkmark          &                        &                        & \checkmark             &            \\
EvoGraphDice~\cite{Boukhelifa2013}	&  &  &  & \checkmark     & \checkmark        & \checkmark        &         \checkmark            &                     &                     & \checkmark             & \checkmark             & \checkmark                         & \checkmark             \\
Dis-Function~\cite{Brown2012}	& & & & \checkmark            &                   & \checkmark        &         \checkmark            &                     & \checkmark          &                        &                        & \checkmark             & \checkmark             \\
UTOPIAN~\cite{Choo2013}	&  &  &  & \checkmark             & \checkmark        &                   &         \checkmark            & \checkmark          &                     &                        &                        & \checkmark             &             
\\\noalign{\global\arrayrulewidth=0.4mm}\hline
\end{tabular}
\caption{Summary of reviewed interactive machine learning systems, characterised by the types of human feedback (implicit, explicit, or both, i.e. mixed), system feedback, the evaluation methods (case study, user study, observational study, survey), and evaluation metrics (objective or subjective ). These systems are ordered in terms of the machine learning tasks they each support: classification, clustering, density estimation or dimensionality reduction.}
\label{reivew-table}
\end{table}

\subsection{Method}
We systematically reviewed papers published between 2012-2017 from the following venues: IEEE VIS, ACM CHI, EG EuroVis, HILDA workshop, and CHI HCML workshop. We downloaded then filtered the proceedings to include papers having the following keywords: ``learn AND algorithm AND interact AND (user OR human OR expert) AND (evaluation OR study OR experiment)''. %This resulted in 1683 papers.
We then drilled down to find papers that describe an actual iML system (as defined in the introduction) with an evaluation section. In this chapter, we focus on studies from the fields of visualization and human-computer interaction. Our hypothesis was that papers from these domains are likely to go beyond algorithm-centered evaluations. In total, we reviewed 19 recent papers (Table~\ref{reivew-table}), from various application domains including multidimensional data\index{multidimensional data} exploration~\cite{Behrisch2014,Boukhelifa2017,Dabek2016,Healey2012,Wenskovitch2017}, data integration~\cite{Azuan2017}, knowledge base construction~\cite{Ehrenberg2016}, text document retrieval~\cite{Heimerl2012}, photo enhancement~\cite{Koyama2016}, audio source separation~\cite{Bryan2014}, social network access control~\cite{Amershi2012}, and category exploration and refinement~\cite{Lin2017}. We examined these evaluations in terms of the machine learning tasks they support, the types of user feedback, the nature of system feedback, and their evaluation methods and metrics.

\subsection{Human Feedback}
Broadly speaking, human feedback to machine learning algorithms can be either \emph{explicit} or \emph{implicit}. The difference between these two mechanisms stems from the field of Information Retrieval (IR). In the case of implicit feedback, humans do not assess relevance for the benefit of the IR system, but rather to fulfill their own task. Besides, they are not necessarily aware that their assessment is being used for relevance feedback~\cite{Kelly2003}. In contrast, for explicit feedback, humans indicate their assessment via a suitable interface, and are aware that their feedback is interpreted for relevance judgment. Whereas implicit feedback is \emph{inferred} from human interactions with the system, explicit feedback is directly provided by humans.

The systems we reviewed either use implicit (7 papers), explicit (8 papers), or mixed (4 papers) human feedback. In the case of mixed feedback, the system tries to infer information from user interactions to complement the explicit feedback.\newline 

\par{\textbf{Implicit Human Feedback}}\newline
Endert et al.~\cite{Endert2012a,Endert2012b} developed \emph{semantic interaction} for visual analytics where the analytical reasoning of the user is inferred from their interactions, which in turn helps steer a dimension reduction model. Their system ForceSpire learns from human input, e.g. moving objects, to improve an underlying model and to produce an improved layout for text documents. Similarly, UTOPIAN~\cite{Choo2013} supports what the authors describe as a ``semantically meaningful set of user interactions'' to improve topic modelling. These interactions include keyword refinement, and topic splitting and merging. Implicit feedback may also be gathered from user interactions with raw data. For example, Azuan et al.~\cite{Azuan2017} developed a tool where manual data corrections, such as adding or removing tuples from a data table, are leveraged to improve data integration and cleaning.

Interactive machine learning systems may infer other types of information such as attribute salience or class membership. Wenskovitch and North implemented Observation-Level Interaction technique (OLI)~\cite{Wenskovitch2017}, where the importance of data attributes is infered from user manipulations of nodes and clusters, and is used to improve a layout algorithm. The ReGroup tool~\cite{Amershi2012} learns from user interactions and faceted search on online social networks to create custom on-demand groups of actors in the network.

In the previous examples, the system learns from individual users. In contrast, Dabek and Caban~\cite{Dabek2016} developed an iML system that learns from crowd interactions with data to generate a user model capable of assisting analysts during data exploration.\newline

\par{\textbf{Explicit Human Feedback}}\newline
Often explicit human feedback is provided through annotations and labels. This feedback can be either binary or graduated. The View Space Explorer~\cite{Behrisch2014} for instance, allows users to choose and annotate relevant or irrelevant example scatter plots. Gao et al.~\cite{Gao2014} proposed an interactive approach to 3D model repository exploration where a human assigns ``like'' or ``dislike'' labels to parts of a model or its entirety. RCLens~\cite{Lin2017} supports user guided exploration of rare categories through labels provided by a human. In a text document retrieval application~\cite{Heimerl2012}, humans decide to accept, reject or label search query results. Similarly, but for a video search system~\cite{Legg2013}, users can either accept or reject sketched query results.

A richer and more nuanced approach to human feedback is proposed by Brown et al. in their Dis-function system~\cite{Brown2012}, where selections of scatterplot points can be dragged and dropped to reflect human understanding of the structure of a text document collection. In this case, the closer the data points in the projected 2D space, the more similar they are. Ehrenberg et al.~\cite{Ehrenberg2016} proposed the ``data programming'' paradigm, where humans encode their domain expertise using simple rules, as opposed to the traditional method of hand-labelling training data. This allows to generate a large amount of noisy training labels, which the machine learning algorithm then tries to de-noise and model. Bryan et al.~\cite{Bryan2014} implemented an audio source separation system where humans annotate data and errors, or directly paint on a time-frequency or spectrogram display. In each of these cases, human feedback and choices are taken into consideration to update a machine learning model.\newline

\par{\textbf{Mixed Human Feedback}}\newline
To guide user exploration of large search spaces, EvoGraphDice~\cite{Boukhelifa2013,Boukhelifa2017} combines explicit human feedback regarding the pertinence of evolved 2D data projections, and an implicit method based on past human interactions with a scatterplot matrix. For the explicit feedback, the user ranks scatterplots from one to five using a slider. The system also infers view relevance by looking at the visual motifs~\cite{Wilkinson2005} in the ranked scatterplots. For example, if the user tends to rank linear point distributions highly, then this motif will be favored to produce the next generation of scatterplots. Importantly, the weights of these feedback channels are set to equal by default, but the user can choose to change the importance of each at any time during the exploration.

Healey and Dennis~\cite{Healey2012} developed interest-driven navigation in visualization, based on both implicit and explicit human feedback. The implicit feedback is gathered from human interactions with the visualization system, and from eye tracking to infer preferences based on where the human is looking. Their argument is that data gathered through implicit feedback is noisy. To overcome this, they built a preference statement interface, where humans provide a subject, a classification, and a certainty. This preference interface allows the human to define rules to identify known elements of interest. 

Another example is the SelPH system~\cite{Koyama2016}, which learns implicitly from a photo editing history, and explicitly from the direct interaction of a human with an optimisation slider. Together, these two feedback channels help to exclude what the authors call the ``uninteresting'' or ``meaningless'' design spaces.

\subsection{System Feedback}
System feedback goes beyond showing the results of the co-operation between the human and the machine. It seeks to inform humans about the state of the machine learning algorithm, and the provenance of system suggestions, especially in the case of implicit user feedback. 

System feedback can be \emph{visual}: 
Boukhelifa et al.~\cite{Boukhelifa2013} used color intensity and a designated flag to visualise the system's interpretation of the mixed user feedback regarding the pertinence of 2D projections. Heimerl et al.~\cite{Heimerl2012} implemented a visual method and text labels to show the classifier's state, and the relevance of the selected documents to a search query. Legg et al.~\cite{Legg2013} visualised the similarity metrics they used to compute a visual search.

System feedback can be \emph{uncertain}: Koyama at al.~\cite{Koyama2016} indicated the system's confidence in the estimation of humans' preferences with respect to color enhancement. Behrisch et al.~\cite{Behrisch2014} provided a feature histogram and an incremental decision tree. These meta visualizations also communicate the classifier's uncertainty. Lin et al.~\cite{Lin2017} showed visualization of rare categories using their ``category view", and a glyph-based visualization to show classification features as well as confidence.

System feedback can be \emph{progressive}: Dabek and Caban~\cite{Dabek2016} discussed the importance of choosing when to propose something to the human. Their approach consisted in providing feedback when the human is in need of guidance. They established a number of rules to detect when this occurs. UTOPIA~\cite{Choo2013} visualises intermediate output even before algorithmic convergence. Ehrenberg et al.~\cite{Ehrenberg2016} showed ``on-the-spot" performance feedback using plots and tables. They claimed that this allows the user to iterate more quickly on system design, and helps navigate the key decision points in their data programming workflow.

For the majority of the iML systems we reviewed, system feedback was provided. It appears that this feedback is an important feature, perhaps because it helps humans better interpret the results, and allows them to correct any mistakes or areas of uncertainty in the inferred user model. The challenge, however, is to find the right level of feedback without having to fully expose the inner workings of the underlying models and their parameters.

\subsection{Evaluation Methods and Metrics}
In total, for the systems we reviewed, there were nine papers with case studies and usage scenarios~\cite{Behrisch2014,Choo2013,Ehrenberg2016,Endert2012a,Gao2014,Healey2012,Legg2013,Lin2017,Wenskovitch2017}, ten user studies~\cite{Amershi2012,Azuan2017,Boukhelifa2017,Brown2012,Bryan2014,Dabek2016,Gao2014,Heimerl2012,Koyama2016,Legg2013} and two observational studies~\cite{Boukhelifa2013,Endert2012b}, in addition to surveys, questionnaires and interviews (seven papers). Although a number of papers included some form of a controlled user study, it was however acknowledged that this type of evaluation is generally difficult to conduct due to the various potential confounding factors such as previous knowledge~\cite{Legg2013}. Indeed, evaluating accuracy of an iML system is not always possible as ground truth does not always exist~\cite{Amershi2012}.\newline

\par{\textbf{Objective Performance Evaluations}}\newline
One way to evaluate how well the human-machine co-operation performs to achieve a task is to compare the iML system with its non-interactive counterpart, i.e. no human feedback, or to an established baseline system. Legg at al.~\cite{Legg2013} conducted a small-scale empirical evaluation with three participants using three metrics inspired from content-based information retrieval: time, precision and recall. The idea was to manually identify five video clips as the ground truth, then to compare an iML video search system with a baseline system (a standard video tool with fast-forward) for a video search task. They found that participants performed better in the iML condition for this task. In a user study with twelve participants, Amerish et al.~\cite{Amershi2012} compared traditional manual search to add people to groups on online social networks (using an alphabetical list or searching by name), to an interactive machine learning approach called ReGroup. They looked at the overall time it took participants to create groups, final group sizes, and speed of selecting group members. Their results show that the traditional method works well for small groups, whereas the iML method works best for larger and more varied groups.

Another way to objectively evaluate the success of the human-machine co-operation is to look at insights. In the context of exploratory data visualization, Endert et al.~\cite{Endert2012a} and Boukhelifa et al.~\cite{Boukhelifa2013} found that with the help of user feedback, their respective iML systems were able to confirm known knowledge and led to new insights.  

Other evaluations in this category compared the iML application with and without system feedback. Dabek et al.~\cite{Dabek2016} proposed a grammar-based approach to model user interactions with data, which is then used to assist other users during data analysis. They conducted a crowdsourced formal evaluation with 300 participants to assess how well their grammar-based model captures user interactions. The task was to explore a census dataset and answer twelve open-ended questions that required looking for combinations of variables and axis ranges using a parallel coordinates visualization. When comparing their tool with and without system feedback, they found that system suggestions significantly improved user performance for all their data analysis tasks, although questions remain with regards to the optimal number of suggestions to display to the user.

A number of studies looked at algorithmic performance when user feedback was implicit versus explicit. Azuan et al.~\cite{Azuan2017} who used a ``pay-as-you-go'' approach to solicit user feedback during data integration and cleaning, compared the two human feedback methods for a data integration task. They found that user performance under the implicit condition was better than for the explicit feedback in terms of number of errors. However, the authors noted some difficulties in separating usability issues related to the explicit feedback interface from the performance results.

Finally, some authors focused on algorithm-centered evaluations, where two or more machine learning methods are compared. For instance, in the context of topic modelling, Choo et al.~\cite{Choo2013} compared latent Dirichlet allocation and non-negative matrix factorisation algorithms, from the practical viewpoints of consistency of multiple runs and empirical convergence. Another example is by Bryan et al.~\cite{Bryan2014} who chose objective separation quality metrics defined by industry standards, as objective measures of algorithmic performance for audio source separation.\newline

\par{\textbf{Subjective Performance Evaluations}}\newline
The subjective evaluations described in Table~\ref{reivew-table} were carried out using surveys, questionnaires,  interviews, and informal user feedback. They included evaluation metrics related to these aspects of user experience: happiness, easiness, quickness, favorite, best helped, satisfaction, task load, trust, confidence in user and system feedback, and distractedness. Moreover, the observational studies~\cite{Boukhelifa2013,Endert2012b} that we reviewed provided rich subjective user feedback on iML system performance. Endert et al.~\cite{Endert2012b} looked at semantic interaction usage, in order to assess whether the latter aids the sensemaking process. They state that one sign of success of iML systems is when humans forget that they are feeding information to an algorithm, and rather focus on ``synthesising information relevant to their task''.

Other evaluations looked at human behavioural variations with regards to different iML interfaces. Amerish et al.~\cite{Amershi2012} compared two interfaces for adding people to online social networks, with and without the interactive component of iML. They looked at behavioural discrepancies in terms of how people used the different interfaces and how they felt. They found that participants were frustrated when model learning was not accurate. Koyama et al.~\cite{Koyama2016} compared their adaptive photo enhancement system with the same tool stripped of advanced capabilities, namely the visual system feedback, the optimisation slider functions, and the ordering of search results in terms of similarity. Because photo enhancement quality can be subjective, performance of the iML system was rated by the study participants. In this case, they were satisfied with the iML system and preferred it over more traditional workflows.\newline

\textbf{In summary, }
There are many aspects of interactive machine learning systems that are being evaluated. Sometimes authors focus on the quality of the user interaction with the iML system (\emph{human-centered evaluations}), or the robustness of the algorithms that are deployed (\emph{algorithm-centered evaluations}), and only in a few cases detailed attention is drawn to the quality of human-machine co-operation and learning. These studies use a variety of evaluation methods, as well as objective and subjective metrics. Perhaps our main observation from this literature review, is that for the majority of the reviewed papers, only a single aspect of the iML system is evaluated. We need more evaluation studies that examine the different aspects of iML sytems, not only as separate components but also from an integrative point of view.

In the next section, we introduce an interactive machine learning system for guided exploratory visualization, and describe our \emph{multi-faceted} evaluation approach to study the effectiveness and usefulness of this tool for end users.

\section{Case Study: Interactive Machine Learning For Guided Visual Exploration}
Exploratory visualization\index{Exploratory visualization} is a dynamic process of discovery that is relatively unpredictable due to the absence of a-priori knowledge of what the user is searching for \cite{Grinstein96}. The focus in this case is on the organisation, testing, developing concepts, finding patterns and definition of assumptions \cite{Grinstein96}. When the search space is large, as is often the case for multi-dimensional datasets, the task of exploring and finding interesting patterns in data becomes tedious. Automatic dimension reduction techniques, such as principle component analysis and multidimensional scaling, reduce the search space, but often are difficult to understand~\cite{Sedlmair2012}, or require the specification of objective criteria to filter views before exploration. Other techniques guide the exploration towards the most interesting areas of the search space based on information learned during the exploration, which appears to be more adapted to the free nature of exploration~\cite{Boukhelifa2017,Brown2012}.

\begin{figure}[h]
  \centering
  \includegraphics[width=.8\linewidth]{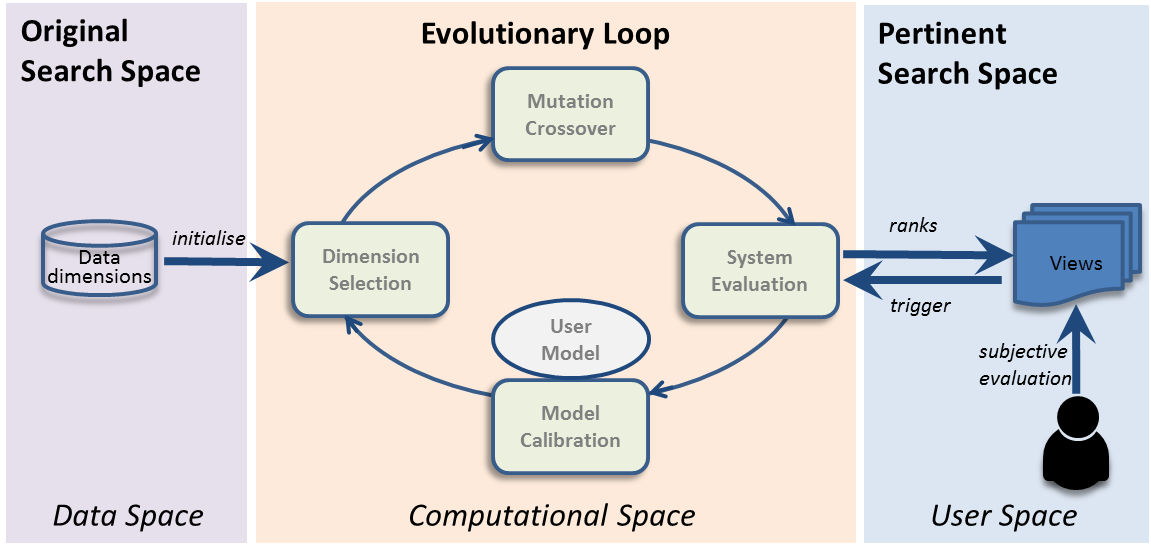}
  \caption{\label{fig:EVE} The Evolutionary Visual Exploration Framework (EVE). Raw data dimensions (from the data space) are fed into an evolutionary loop in order to progressively evolve new interesting views to the user. The criteria for deciding on the pertinence of the new views is specified through a combination of automatically calculated metrics (from the computational space) and user interactions (at the user space).}
\end{figure}

In our previous work on guided exploratory visualization \cite{Boukhelifa2017,Boukhelifa2015,Boukhelifa2013,Ticona2013,Ticona2012}, we tried to address the problem of how to efficiently explore multidimensional datasets characterised by a large number of projections. We proposed a framework for Evolutionary Visual Exploration (EVE, Figure~\ref{fig:EVE}) that combines visual analytics\index{visual analytics} with stochastic optimisation by means of an Interactive Evolutionary Algorithm (IEA). Our goal was to guide users to interesting projections, where the notion of ``interestingness" is defined \emph{implicitly} by automatic indicators such as the amount of visual pattern in the two-dimensional views visited by the user, and \emph{explicitly} via subjective human assessment.

In this section, we report on our experience in building and evaluating an interactive machine learning system called EvoGraphDice (Figure~\ref{fig:evographdice}) using the EVE framework. We note that existing evaluations of interactive evolutionary systems tend to be algorithm-centered. Through this case study, we argue for a \emph{multi-faceted} evaluation approach that takes into account all components of an iML system. Similar recommendations can be found for evaluating interactive visualization systems. For example, Carpendale~\cite{Carpendale2008} advocates for adopting a variety of evaluative methodologies that together may start to approach the kind of answers sought.

\subsection{Background on Interactive Evolutionary Computation IEC}
There are many machine learning approaches, including artificial neural networks, support vector machines and Bayesian networks. Moreover, many machine learning problems can be modelled as optimisation problems where the aim is to find a trade-off between an adequate representation of the training set and a generalisation capability on unknown samples. In contrast to traditional local optimisation methods, Evolutionary Algorithms (EAs) have been widely used as a successful stochastic optimisation tool in the field of machine learning in the recent years~\cite{Stanley2002}. In this sense, machine learning and the field of Evolutionary Computation\index{Evolutionary Computation} (EC), that encompasses EAs, are tightly coupled.

\begin{figure}
  \begin{center}
    \centering
    \includegraphics[width=.5\linewidth]{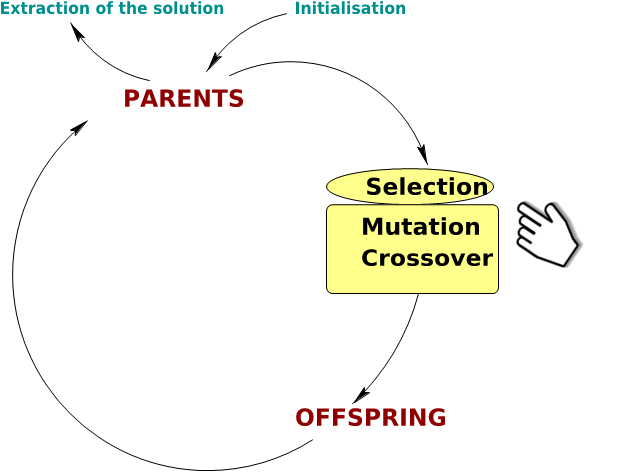}
    \caption{The evolutionary loop: user interactions can occur at any stage including the selection and evaluation of individuals and the genetic operators.}~\label{fig:evoloop}
  \end{center}
\end{figure}

Evolutionary Algorithms (EAs) are stochastic optimisation heuristics that copy, in a very abstract manner, the principles of natural evolution that let a population of individuals be adapted to its environment~\cite{Goldberg1989}. They have the major advantage over other optimisation techniques of making only few assumptions on the function to be optimised. An EA considers populations of potential solutions exactly like a natural population of individuals that live, fight, and reproduce, but the natural environment pressure is replaced by an ``optimisation'' pressure. In this way, individuals that reproduce are the best ones with respect to the problem to be solved.
Reproduction (see Figure~\ref{fig:evoloop}) consists of generating new solutions via variation schemes (the genetic operators), that, by analogy with nature, are called mutation if they involve one individual, or crossover if they involve two parent solutions. A \emph{fitness function}, computed for each individual, is used to drive the selection process, and is thus optimised by the EA. Evolutionary optimisation techniques are particularly efficient to address complex problems (irregular, discontinuous) where classical deterministic methods fail~\cite{banzhaf1997,poli1997}, but they can also deal with varying environments~\cite{jin2005}, or non computable quantities~\cite{takagi2001}.

Interactive Evolutionary Computation\index{Interactive Evolutionary Computation} (IEC) describes evolutionary computational models where humans, via suitable user interfaces, play an active role, \emph{implicitly} or \emph{explicitly}, in evaluating the outputs evolved by the evolutionary computation (Figure~\ref{fig:evoloop}). IEC lends itself very well to art applications such as for melody or graphic art generation where creativity is essential, due to the subjective nature of the fitness evaluation function. For scientific and engineering applications, IEC is interesting when the exact form of a more generalised fitness function is not known or is difficult to compute, say for producing a visual pattern that would interest a human observer. Here, the human visual system, together with their emotional and psychological responses are far superior than any automatic pattern detection or learning algorithm.

Whereas current IEC research has focused on improving the robustness of the underlying algorithms, much work is still needed to tackle human-factors in systems where adaptation between users and systems is likely to occur~\cite{Mackay2000}.

\subsection{The Visible and Hidden Roles of Humans in IEC}
The role of humans in IEC can be characterised by the evolutionary component at which they operate, namely: initialisation, evolution, selection, genetic operators, constraints, local optimisation, genome structure variation, and parameters tuning. This may or may not be desirable from a usability perspective, especially for non-technical users. The general approach when humans are involved, especially for parameter tuning, is mostly by trial-and-error and by reducing the number of parameters. Such tasks are often visible, in that they are facilitated by the user interface. However, there exists a hidden role of humans in IEC that has often been neglected. Algorithm and system designers play a central role in deciding the details of the fitness function to be optimised and in setting the default values of system parameters, and thus contributing to the ``black-box'' effect of IEC systems. Such tasks are influenced by the designer's previous experience and end-user task requirements.

Besides this hidden role in the design stage, there is a major impact of the ``human-in-the-loop'' on the IEC. This problem is known as the ``user bottleneck'', i.e. human fatigue due to the fact that the human and the machine do not live and react at the same rate. Various solutions have been considered in order to avoid systematic and repetitive or tedious interactions, and the authors themselves have considered several of them, such as: (i) reducing the size of the population and the number of generations; (ii) choosing specific models to constrain the exploration in a-priori ``interesting'' areas of the search space; and (iii) performing an automatic learning (based on a limited number of characteristic quantities) in order to assist the user and only present interesting individuals of the population, with respect to previous votes or feedback from the user. These solutions require considerable computational effort. A different approach and new ideas to tackle the same issue could come from Human Computer Interaction (HCI) and usability research, as discussed later on in this chapter.

\subsection{EvoGraphDice Prototype}
\begin{figure}
  \begin{center}
    \centering
    \includegraphics[width=\linewidth]{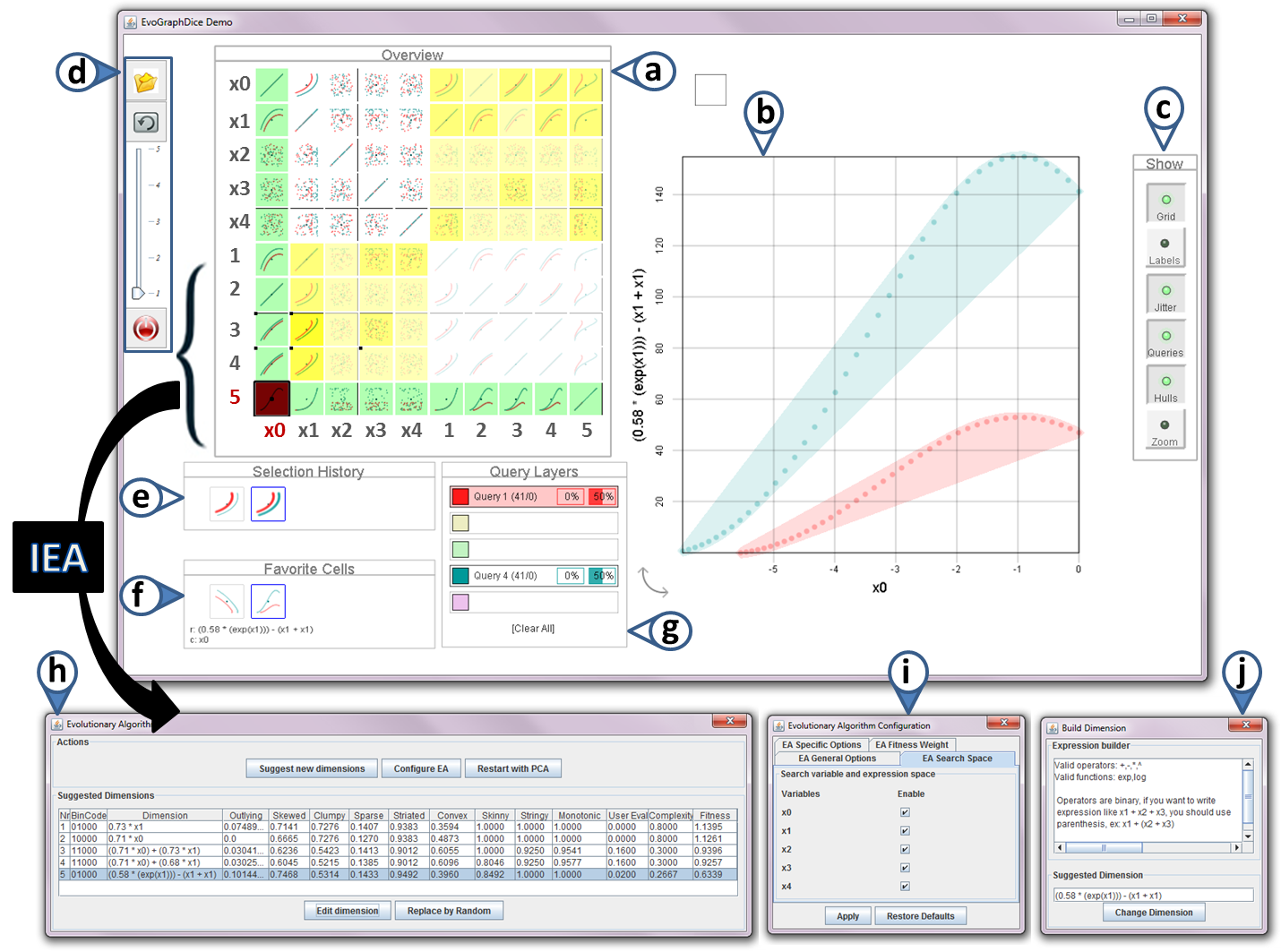}
    \caption{EvoGraphDice prototype showing an exploration session of a synthetic dataset. Widgets: (a) an overview scatterplot matrix showing the original data set of 5 dimensions (x0..x4) and the new dimensions (1..5) as suggested by the evolutionary algorithm. (b) main plot view. (c) tool bar for main plot view. (d) a tool bar with (top to bottom) “favorite” toggle button, “evolve” button , a slider to evaluate cells and a restart (PCA) button. (e) the selection history tool. (f) the favorite cells window. (g) the selection query window. (h) IEA main control window. (i) window to limit the search space. (j) dimension editor operators.}~\label{fig:evographdice}
  \end{center}
\end{figure}

EvoGraphDice~\cite{Boukhelifa2017,Boukhelifa2015,Boukhelifa2013,Ticona2012} was designed to aid the exploration of multidimensional datasets characterised by a large space of 2D projections (Figure~\ref{fig:evographdice}). Starting from dimensions whose values are automatically calculated by a Principle Component Analysis (PCA), an IEA progressively builds non-trivial viewpoints in the form of linear and non-linear dimension combinations, to help users discover new interesting views and relationships in their data. The criteria for evolving new dimensions is not known a-priori and is partially specified by the user via an interactive interface. Pertinence of views is modelled using a fitness function that plays the role of a predictor: (i) users select views with meaningful or interesting visual patterns and provide a satisfaction score; (ii) the system  calibrates the fitness function optimised by the evolutionary algorithm to incorporate user's input, and then calculates new views. A learning algorithm was implemented to provide pertinent projections to the user based on their past interactions.\newline 

\subsection{Multi-Faceted Evaluation of EvoGraphDice}
We evaluated EvoGraphDice quantitatively and qualitatively following a mixed-approach, where on the one hand we analysed the computational behaviour of the system (algorithm-centered approach), and on the other hand we observed the utility and effectiveness of the system for the end-user (human-centered approach).

\subsubsection{Quantitative Evaluation}
For this study~\cite{Boukhelifa2017}, we synthesised a 5D dataset with an embedded curvilinear relationship between two dimensions and noise for the rest of the dimensions. The task was to find a data projection that shows a derived visual pattern. We logged user interactions with the tool and the state of the system at each algorithm iteration. For log data analysis, we used both statistical and exploratory visualization techniques.\newline

\textbf{Algorithm-Centered Evaluation}\newline
This evaluation focused on two aspects of our iML system: the \emph{robustness} of the underlying algorithm, and the \emph{quality of machine learning}. To study robustness, we conducted two types of analyses: (a) \emph{convergence analysis} to assess the algorithm’s ability to steer the exploration toward a focused area of the search space, and (b) \emph{diversity analysis} to assess the richness and variability of solutions provided by the algorithm. These two analyses are relevant because they relate to two important mechanisms in evolutionary algorithms, \emph{exploitation} and \emph{exploration}~\cite{banzhaf1997}, where on the one hand users want to visit new regions of the search space, and on the other hand they also want to explore solutions close to one region of the search space. In terms of objective metrics, we used the number of generations and task outcome to measure algorithmic performance, and mean visual{} pattern differences (using scagnostics~\cite{Wilkinson2005}) to assess diversity. To evaluate the quality of learning, we used the rate of concordance between user evaluation scores, and the ``predicted'' values as calculated by the algorithm.

Our analysis showed that on average the interactive evolutionary algorithm followed the order of user ranking of scatterplots fairly consistently, even though users seemed to take different search and evaluation strategies. For example, some participants tended to lump evaluation scores to fewer levels, others used the five provided score levels, whereas the rest alternated between the two strategies at different stages of the exploration. Moreover, these results indicated a possible link between user evaluation strategy, and outcome of exploration and speed of convergence, where users taking a more consistent approach converged more quickly. The diversity analysis showed that, in terms of visual pattern, the IEA provided more diverse solutions at the beginning of the exploration session before slowly converging to a more focused search space.\newline

\textbf{Human-Centered Evaluation}\newline
The user-centered evaluation of EvoGraphDice focused on two different aspects related to human interactions with the iML system. First we performed a \emph{user strategy analysis} to understand the different approaches users took to solve a data exploration task. The evaluation metrics we used here were the type of searched visual pattern, and stability of the exploration strategy. Second, we looked at \emph{user focus} to highlight hot spots in the user interface and assess user evaluation strategies. In this case, our evaluation metrics were related to the user view visitation and evaluation patterns.

In terms of results, the user strategies analysis showed that EvoGraphDice allows for different types of exploration strategies that appear to be relevant for the study task. In the case of a two-curve separation task, these strategies centered around three dominant types of scagnostics: skinny, convex and sparse. We also found that the stability of the exploration strategy may be an important factor for determining the outcome of the exploration task and the speed of convergence, since successful exploration sessions had a more consistent strategy when compared to the unsuccessful ones, and they converged more quickly on average. 

From the user visitation and evaluation analyses, we found that users were more likely to visit scatterplots showing dimensions relevant to their task. Moreover, these plots were on average ranked highly by the user. Since for this game task, the main dimensions relevant to the task appeared on the top left side of the proposed cells, users intuitively started navigating that way. What we saw in these results was probably a mixture of task-relevance and intuitive-navigation, as the relevant original dimensions are placed in a prominent position in the matrix.\newline

\subsubsection{Qualitative Evaluation}
To assess the usability and utility of EVE, we conducted another user study~\cite{Boukhelifa2013} where we tried to answer these three questions: is our tool understandable and can it be learnt; are experts able to confirm known insights in their data; and are they able to discover new insight and generate new hypotheses. We designed three tasks: (a) a game-task (similar to the task in the quantitative evaluation above) with varying levels of difficulty to assess participants abilities to operate the tool; (b) we asked participants to show in the tool what they already know about their data; and (c) to explore their data in light of a hypothesis or research question that they prepared. This sequence of tasks assured that experts became familiar with the tool, and understood how to concretely leverage it by looking for known facts, before looking for new insights. This evaluation approach sits between an observational study and an insight-based evaluation, such as the one proposed by Saraiya et al.~\cite{saraiya2005}.

The study led to interesting findings such as the ability of our tool to support experts in better formulating their research questions and building new hypotheses. For insight evaluation studies such as ours, reproducing the actual findings across subjects is not possible as each participant provided their own dataset and research questions. However, reproducing testing methodologies and coding for the analysis is. Although we run multiple field studies with domain experts from different domains, with sessions that were internally very different, the high-level tasks, their order and the insight-based coding were common. Training expert users on simple specific tasks that are not necessarily ``theirs" also seemed to help experts become confident with the system, but of course comes at a time cost.

\section{Discussion}
We conducted qualitative and quantitative user studies to evaluate EVE which helped us validate our framework of guided visual exploration. While
the observational study showed that using EVE, domain experts were able to formulate interesting hypothesis and reach new insights when exploring freely, the quantitative evaluation indicated that users, guided by the interactive evolutionary algorithm, are able to converge quickly to an interesting view of their data when a clear task is specified. Importantly, the quantitative study allowed us to accurately describe the relationship between user behaviour and algorithm’s response.

Besides interactive machine learning, guided visualization systems such as EVE fall under the wider arena of knowledge-assisted visualization and mixed-initiative systems \cite{Horvitz1999}. In such cases, where the system is learning, it is crucial that users understand what the system is proposing or why changes are happening. Thus, when evaluating iML systems with users, we need to specifically test if the automatic state changes and their provenance are understood. It would be interesting, for example, to also consider evolving or progressive revealing of the provenance of system suggestions. This way, as the user becomes more expert, more aspects of the underlying mechanics are revealed. When creativity and serendipity are important aspects, as it is the case in artistic domains and data exploration, new evaluation methodologies are required.

Research from the field of mixed initiative systems describes a set of design principles that try to address systematic problems with the use of automatic services within direct manipulation interfaces. These principles include considering uncertainty about a user's goal, transparency, and considering the status of users' attention \cite{Horvitz1999}. We can be inspired by the extensive experience and past work from HCI, to also consider how user behaviour can in turn adapt to fit our systems \cite{Mackay2000}.

During the design, development and evaluation of EVE, we worked with domain experts at different levels. For the observational study, we worked with data experts from various disciplines, which allowed us to asses the usefulness, usability and effectiveness of our system in different contexts. In particular, we largely benefited from having one domain expert as part of the design and evaluation team. This expert explored multidimensional datasets as part of her daily work, using both algorithmic and visual tools. Involving end-users in the design team is a long-time tradition in the field of HCI as part of the user-centered design methodology. This is a recommendation we should consider more, both as a design and as a system validation approach. While HCI researchers acknowledge the challenges of forming partnerships with domain experts, their past experience (e.g. \cite{Chilana2010}) can inform us on how to proceed with the evaluation of iML systems.

\section{Research Prospects}
We report on observations and lessons learnt from working with application users both for the design and the evaluation of our interactive machine learning system, as well as the results of experimental analyses. We discuss these below as research opportunities aiming to facilitate and support the different roles humans play in iML, i.e. in the design, interaction and evaluation of these systems.\newline

\par \textbf{Human-Centered Design:} during the design, development and evaluation of many of our tools, we worked with domain experts at different levels. For EvoGraphDice, for instance, we largely benefited from having a domain expert as part of the design and evaluation team. However, this was carried out in an informal way. Involving end-users in the design team is a long-time tradition in the field of HCI as part of the user-centered design methodology. Participatory design, for instance, could be conducted with iML end-users to incorporate their expertise in the design of, for example, learning algorithms and user models. This is a recommendation we should consider in a more systematic way, both as a design and as a system validation approach.\newline

\par \textbf{Interaction and Visualization:} often the solutions proposed by the iML systems are puzzling to end-users. This is because the inner workings of machine learning algorithms, and the user exploration and feedback strategies that led to system suggestions are often not available to the user. This ``black-box'' effect is challenging to address as there is a fine balance to find between the richness of a transparent interface and the simplicity of a more obscure one. Finding the tipping point requires an understanding of evolving user expertise in manipulating the system, and the task requirements. Whereas HCI and user-centered design can help elicit these requirements and tailor tools to user needs over time, visualization techniques can make the provenance of views and the system status more accessible.

At the interaction level, HCI can contribute techniques to capture rich user feedback without straining the user, that are either implicit (e.g. using eye-tracking); or explicit such as using simple gestures or interactions mediated by tangible objects to indicate user subjective assessment of a given solution. Here, our recommendation is to investigate rich and varied interaction techniques to facilitate user feedback, and to develop robust user models that try to learn from the provided input.\newline

\par \textbf{Multifaceted Evaluation:} the evaluation of iML systems remains a difficult task as often the system adapts to user preferences but also the user interprets and adapts to system feedback. Getting a clear understanding of the subtle mechanisms of this co-adaptation~\cite{Mackay2000}, especially in the presence of different types and sources of uncertainty~\cite{Boukhelifa2017b}, is challenging and requires to consider evaluation criteria other than speed of algorithm convergence and the usability of the interface.

In the context of exploration, both for scientific and artistic applications, creativity is sought and can be characterised by lateral thinking, surprising  findings, and the way users learn how to operate the interactive system and construct their own way to use it. For IEC, our observation is that augmented creativity can be achieved with the right balance between randomness and user-guided search. What is important to consider for evaluating iML systems in the context of creativity, are the exploration components. Our recommendation with this respect is two-fold: first, to work towards creating tools that support creativity (something that the HCI community is already looking into~\cite{Cherry2014}); and second, to investigate objective and subjective metrics to study creativity within iML (e.g. to identify impacting factors such as the optimisation constraints, user engagement and the presence or absence of direct manipulation). Some of these measures may only be identifiable through longitudinal observations of this co-adaptation process.

\section{Conclusion}
User-driven machine learning processes such as the ones described in this chapter, rely on systems that adapt their behaviour based on user feedback, while users themselves adapt their goals and strategies based on the solutions proposed by the system. In this chapter, we focused on the evaluation of interactive machine learning systems, drawing from related work, and our own experience in developing and evaluating such systems. We showed through a focused literature review that despite the multifaceted nature of iML systems, current evaluations tend to focus on single isolated components such as the robustness of the algorithm, or the utility of the interface. Through a visual analytics case study, we showed how coupling algorithm-centered and user-centered evaluations can bring forth insights on the underlying co-operation and co-adaptation mechanisms between the algorithm and the human. Interactive machine learning presents interesting challenges and prospects to conduct future research not only in terms of designing robust algorithms and interaction techniques, but also in terms of coherent evaluation methodologies.

\bibliographystyle{plain}
\bibliography{eval-bib}

\end{document}